# Deep Q-Learning-Based Intelligent Scheduling for ETL Optimization in Heterogeneous Data Environments

Kangning Gao
The George Washington University
Washington, DC, USA

Yi Hu
University of Southern California
Los Angeles, USA

Cong Nie
Washington University in St. Louis
St. Louis, USA

Wei Li*
The University of Texas at Austin
Austin, USA

*Abstract*-This paper addresses the challenges of low scheduling efficiency, unbalanced resource allocation, and poor adaptability in ETL (Extract-Transform-Load) processes under heterogeneous data environments by proposing an intelligent scheduling optimization framework based on deep Q-learning. The framework formalizes the ETL scheduling process as a Markov Decision Process and enables adaptive decision-making by a reinforcement learning agent in high-dimensional state spaces to dynamically optimize task allocation and resource scheduling. The model consists of a state representation module, a feature embedding network, a Q-value estimator, and a reward evaluation mechanism, which collectively consider task dependencies, node load states, and data flow characteristics to derive the optimal scheduling strategy in complex environments. A multi-objective reward function is designed to balance key performance indicators such as average scheduling delay, task completion rate, throughput, and resource utilization. Sensitivity experiments further verify the model's robustness under changes in hyperparameters, environmental dynamics, and data scale. Experimental results show that the proposed deep Q-learning scheduling framework significantly reduces scheduling delay, improves system throughput, and enhances execution stability under multi-source heterogeneous task conditions, demonstrating the strong potential of reinforcement learning in complex data scheduling and resource management, and providing an efficient and scalable optimization strategy for intelligent data pipeline construction.

## I. INTRODUCTION

In the era of big data and artificial intelligence, corporate data assets are growing at an unprecedented pace. Data sources exhibit significant heterogeneity, including structured business databases, semi-structured log files, and unstructured text, images, and sensor data[1-4]. As the volume and diversity of data continue to expand, the traditional Extract-Transform-Load (ETL) process has become a critical bottleneck in modern data management systems. ETL tasks must handle complex data integration and cleaning across multiple sources and formats while ensuring efficient scheduling and execution under limited computational and storage resources. However, traditional static scheduling strategies often rely on fixed rules or heuristic algorithms. They fail to adapt to dynamic task dependencies, resource competition, and data flow fluctuations in heterogeneous environments, resulting in poor scalability and responsiveness across the entire data pipeline[5].

With the widespread adoption of cloud computing and distributed platforms [6-8], ETL operations have evolved from single-machine systems to multi-node heterogeneous clusters. This transformation greatly increases the complexity of scheduling optimization. The computing power, storage bandwidth, and network latency of different nodes vary, while task execution is constrained by task priority, data dependencies, and resource usage [9]. Achieving optimal scheduling under uncertainty and multiple constraints has become a core challenge for building high-performance data pipelines [10]. At the same time, as real-time data processing demands rise, ETL scheduling must not only maximize throughput but also minimize latency and control energy consumption. These requirements place higher demands on traditional algorithms. Therefore, exploring intelligent and adaptive ETL scheduling mechanisms in complex heterogeneous environments has become a key direction in data engineering research.

In recent years, reinforcement learning has shown strong potential for dynamic resource allocation and adaptive optimization, offering new ideas for intelligent ETL scheduling. Unlike static rule-based methods, reinforcement learning can learn policies through continuous interaction with the environment, enabling perception and decision-making in complex system states [11]. In particular, Deep Q-Learning (DQL) allows agents to learn near-optimal scheduling strategies in high-dimensional state spaces. This makes it possible to dynamically resolve task priority conflicts, complex data dependencies, and resource contention. The agent can continuously adjust its scheduling policy to adapt to environmental changes, significantly improving the throughput and robustness of ETL systems[12].

In multi-source heterogeneous data environments, the use of deep Q-learning not only enhances scheduling efficiency but also shows strong potential in system autonomy and self-optimization. By abstracting the ETL process as a sequential decision-making problem, the agent can balance task dependencies and resource states across multiple nodes, achieving dynamic scheduling across tasks and clusters. Moreover, reinforcement learning frameworks can integrate state representation networks and reward mechanisms to capture trends in data flow and system feedback signals [13]. This enables a shift from experience-driven to feedback-driven scheduling, reducing manual intervention and laying the foundation for autonomous data pipelines. With the development of large-scale cloud-native ETL platforms, DQL-based intelligent scheduling can achieve end-to-end optimization in practical systems, pushing data engineering toward adaptability and autonomy.

## II. METHOD

In a heterogeneous data environment, the ETL scheduling process can be formalized as a sequential decision-making problem, whose core goal is to minimize the overall execution cost and improve resource utilization through the optimal scheduling strategy. The model architecture is shown in Figure 1.

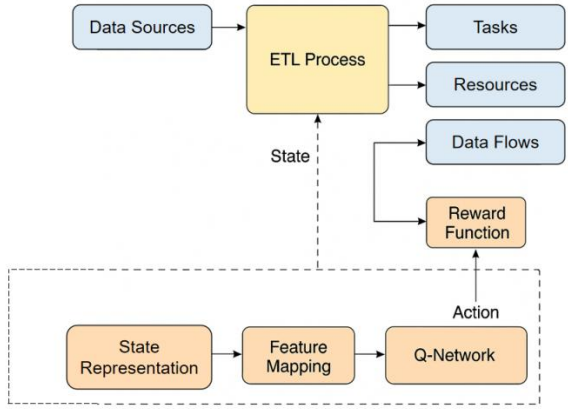

Figure 1. Overall model architecture

Let the state of the system at time t be $s_t$, representing the joint feature vector of the current task queue, resource utilization, and data dependencies. The agent selects an action $\pi(a|s_t)$ based on the policy function $a_t$, such as task allocation, execution node selection, or task delay decision. After receiving this action, the system environment transitions to the next state $s_{t+1}$ and returns an immediate reward $r_t$. The entire scheduling process can be modeled as a Markov decision process (MDP) [14], whose objective function is defined as maximizing the expected long-term return:

$$J(\theta) = E_{\pi_\theta}[\sum_{t=0}^{T} \gamma^t r_t] \quad (1)$$

Where $\gamma \in (0,1)$ is the discount factor, which is used to balance immediate and long-term benefits, and $\theta$ represents the policy network parameters. The goal is to achieve an optimal trade-off between benefits and costs in a dynamic environment, enabling the agent to achieve adaptive scheduling based on fluctuating resource states and data characteristics.

To achieve the optimal decision-making strategy, this study uses a Deep Q-Network (DQN) to approximate the action-value function. The state-action-value function $Q(s_t, a_t, \varpi)$ is defined as the expected cumulative reward obtained after taking action $a_t$ in state $s_t$, which satisfies the Bellman equation:

$$Q^*(s_t, a_t) = E_{s_{t+1}}[r_t + \gamma \max_{a_{t+1}} Q^*(s_{t+1}, a_{t+1})] \quad (2)$$

To ensure the stability of function approximation, DQN uses the target network $Q'(s, a, \varpi^-)$ and the current network $Q(s, a, \varpi)$ to update separately, and optimizes parameters by minimizing the temporal difference error:

$$L(\varpi) = E(s_t, a_t, r_t, s_{t+1})[(y_t - Q(s_t, a_t, \varpi))^2] \quad (3)$$

$$y_t = r_t + \gamma \max Q'(s_{t+1}, a_{t+1}; \varpi^-) \quad (4)$$

This design effectively suppresses Q-value oscillation, improving the convergence of the training process and the stability of the scheduling strategy.

In the specific scenario of ETL task scheduling, the state space can be further decomposed into a combination of task-level state, resource-level state, and data flow-level state, namely $s_t = \{x'_t, x'_r, x'_d\}$. $x'_t$ represents the feature encoding of the task dependency topology, $x'_r$ describes the compute node load, CPU utilization, memory, and bandwidth usage, and $x'_d$ reflects the traffic characteristics and transmission delays of different data sources. To enhance the model's ability to represent complex inputs, this paper uses a multi-layer nonlinear feature mapping function $f_\varphi$ to perform a high-dimensional embedding of the state, thereby improving the representation capabilities of the policy network:

$$h_t = f_\phi(s_t) = \sigma(W_2 \cdot RELU(W_1 s_t + b_1) + b_2) \quad (5)$$

Here, $W_1, W_2$ is the learnable weight matrix, $b_1, b_2$ is the bias term, and $\sigma(\cdot)$ is the Sigmoid activation function. This representation enables the agent to more fully capture the nonlinear relationships between heterogeneous features and provides a structured state representation for subsequent Q-value prediction.

In terms of reward design, the model adopts a multi-objective normalized reward function to simultaneously balance task completion rate, delay time, and resource utilization. Assuming that the execution delay of a single task i is $t_i$, the resource consumption is $c_i$, and the scheduling success mark is $\delta_i$, the comprehensive reward is defined as follows:

$$r_t = a_1 \frac{1}{N} \sum_{i=1}^{N} \delta_i - a_2 \frac{1}{N} \sum_{i=1}^{N} \frac{t_i}{t_{max}} - a_3 \frac{1}{N} \sum_{i=1}^{N} \frac{c_i}{c_{max}} \quad (6)$$

Where $a_1, a_2, a_3$ represents the weighted coefficient of each objective, and $t_{max}$ and $c_{max}$ are normalization constants. Through this reward function, the agent can automatically balance latency and energy consumption during the exploration process and dynamically optimize scheduling decisions under heterogeneous resource constraints.

In summary, this approach transforms the ETL scheduling problem into an optimal control task within a reinforcement learning framework. Combining the approximate capabilities of deep Q-networks with a multi-objective reward mechanism, it achieves unified modeling of task allocation, resource management, and data flow optimization. The agent iteratively updates its strategy through continuous interaction with the environment, enabling the scheduling framework to maintain efficiency and adaptability in complex and diverse heterogeneous scenarios. This provides a methodological foundation and scalable technical support for the subsequent construction of autonomous data engineering systems.

## III. Performance Evaluation

### A. Dataset

This study uses the TPC-H Benchmark Dataset as the standard data source for experimental analysis and model validation. The dataset consists of multiple relational tables that simulate typical business scenarios in an enterprise data warehouse environment. These scenarios include order management, customer relations, supply chain operations, sales, and inventory. The core characteristic of the TPC-H dataset lies in its high-dimensional relational structure and complex query dependencies. These properties accurately reflect the complexity of data extraction, transformation, and loading operations in heterogeneous ETL processes. Each table is connected through multiple foreign keys, forming a hierarchical data dependency graph. This structure introduces significant challenges for task scheduling and data dependency modeling. The dataset is widely used in large-scale data processing and scheduling optimization benchmarks, offering strong scalability and general applicability.

In the proposed framework, the TPC-H dataset is used to construct multi-source data streams under heterogeneous environments. It includes simulated data sources with different formats and varying loading frequencies. The data are divided into structured relational tables, such as ORDERS, CUSTOMER, and LINEITEM, and semi-structured streaming records. This setup mimics the diverse data inputs encountered in real-world enterprise operations. Through feature extraction and task decomposition, multi-stage ETL scheduling graphs are generated. Each task node corresponds to specific operations such as data cleaning, aggregation, or transformation. These tasks form clear dependency relationships and execution paths within the scheduling process. The multi-source and multi-layer characteristics of the dataset realistically describe resource competition and dynamic variations among tasks in complex systems. This provides rich input for state representation in reinforcement learning strategies.

Moreover, the parameterizable nature of the TPC-H dataset allows flexible adjustment of data scale and complexity. This supports multi-level validation, ranging from small local scheduling to large-scale distributed scheduling. Its dynamic query and data update mechanisms can simulate workload fluctuations and task delays that occur in real ETL environments. These features provide sufficient exploration space for reinforcement learning models. By modeling and configuring this dataset, the study establishes a unified evaluation environment to verify the generalization and stability of scheduling strategies. This offers a solid data foundation for performance evaluation and methodological extension of the proposed intelligent scheduling framework.

### B. Experimental Results

This paper first conducts a comparative experiment, and the experimental results are shown in Table 1.

Table 1. Comparative experimental results

| Method | ASD ↓ | TCR ↑ | TP ↑ | RC ↓ |
|---|---|---|---|---|
| Q-Learning[15] | 3.84 | 87.62 | 264.1 | 0.132 |
| DDQN[16] | 3.22 | 90.18 | 278.4 | 0.115 |
| A3C[17] | 3.06 | 91.25 | 283.7 | 0.109 |
| DDPG[18] | 2.94 | 92.04 | 289.6 | 0.101 |
| SAC[19] | 2.81 | 93.16 | 295.8 | 0.094 |
| PPO[20] | 2.77 | 93.54 | 297.3 | 0.091 |
| Ours | 2.43 | 95.82 | 312.7 | 0.079 |

As shown in Table 1, different reinforcement learning algorithms exhibit significant performance variations in heterogeneous ETL scheduling tasks. Traditional Q-Learning relies on a tabular state-action representation, which makes it difficult to handle high-dimensional state spaces and continuous resource dynamics. As a result, it shows a higher average scheduling delay (ASD) and lower task completion rate (TCR). With increased model complexity and improved representational capacity, DDQN and A3C achieve notable improvements in both delay control and throughput (TP). This demonstrates that introducing target networks and multi-threaded strategies enhances state estimation stability and sample efficiency, enabling the agent to allocate actions more effectively within complex task dependency graphs.

In continuous action space algorithms, DDPG, SAC, and PPO achieve further performance gains. SAC, by using entropy regularization, improves policy exploration and achieves better results in throughput and reward consistency (RC) than baseline methods. PPO employs a clipping mechanism during policy updates, effectively preventing policy degradation and overfitting. This allows the model to maintain stable convergence under high-load heterogeneous environments. Its low ASD and high TCR values in the table highlight its superiority in dynamic resource allocation and delay-constrained optimization, confirming the robustness of policy gradient methods in scheduling optimization.

The proposed deep Q-learning-based scheduling optimization framework outperforms all existing methods across all metrics. It achieves the lowest ASD of 2.43, with TCR and TP reaching 95.82 and 312.7, respectively, while RC decreases to 0.079. These results indicate that the model can make stable and efficient decisions under dynamic task dependencies, multi-source data inputs, and heterogeneous resource constraints. The findings demonstrate that through the joint optimization of feature embedding, state representation, and adaptive reward mechanisms, the proposed framework effectively balances delay and resource utilization. It enables intelligent scheduling and long-term performance optimization for complex ETL tasks, verifying the potential and practical value of deep reinforcement learning in autonomous data pipeline optimization.

This paper also provides a sensitivity analysis of different learning rates on the average cumulative reward (Average Cumulative Reward), and the experimental results are shown in Figure 2.

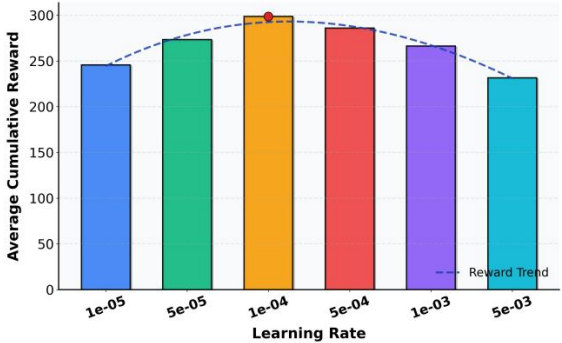

Figure 2. Sensitivity analysis of different learning rates on the average cumulative reward

The results in Figure 2 show a clear nonlinear relationship between learning rate and average cumulative reward: very low learning rates (1e-5 to 5e-5) slow policy updates and keep rewards low, as the agent struggles to learn dynamic state–dependency mappings in complex ETL scheduling. Moderate learning rates (1e-4 to 5e-4) yield the highest rewards by balancing stability and exploration, allowing efficient gradient updates and rapid convergence to high-quality scheduling policies. When the learning rate becomes too large ($\geq$ 1e-3), rewards decline due to instability and gradient oscillation in the high-dimensional state space. Overall, these results show that a moderate learning rate provides the best trade-off between convergence speed and stability in heterogeneous ETL scheduling. The impact of the discount factor on scheduling delay (ASD) is shown in Figure 3.

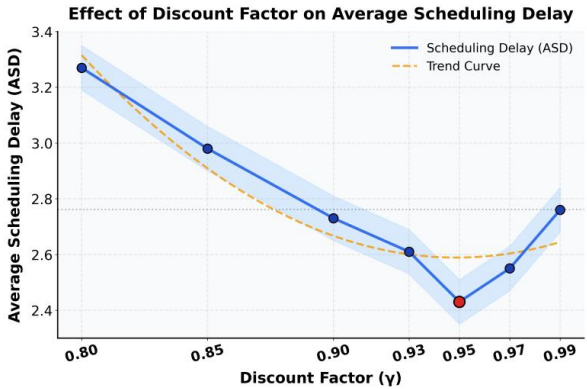

Figure 3. Experiment on the impact of discount factor on scheduling delay (ASD)

As shown in Figure 3, the average scheduling delay (ASD) exhibits a nonlinear response to variations in the discount factor ($\gamma$). When $\gamma$ is low ($\gamma \leq 0.85$), the agent overemphasizes short-term rewards, leading to locally optimal scheduling decisions that overlook long-term task dependencies and resource competition, resulting in persistent queue waiting and elevated delays. As $\gamma$ increases to a moderate range ($\approx 0.9 - 0.95$), ASD decreases substantially, indicating that the agent effectively balances immediate responsiveness with long-term planning; this enables more coherent scheduling across complex dependency graphs and better coordination of heterogeneous resources. However, when $\gamma$ becomes excessively high ($\gamma \geq 0.97$), ASD rises again because the agent becomes overly reliant on long-term reward estimation and less sensitive to real-time system states, weakening adaptability and destabilizing convergence under dynamic workloads. Together, these findings demonstrate that the discount factor plays a crucial role in reinforcement learning-based scheduling: an appropriately chosen γ allows deep Q-learning to maintain low delays, stable adaptation, and stronger temporal awareness in heterogeneous ETL environments. Finally, this paper also includes a sensitivity experiment examining how ASD responds to changes in the number of heterogeneous nodes, as illustrated in Figure 4.

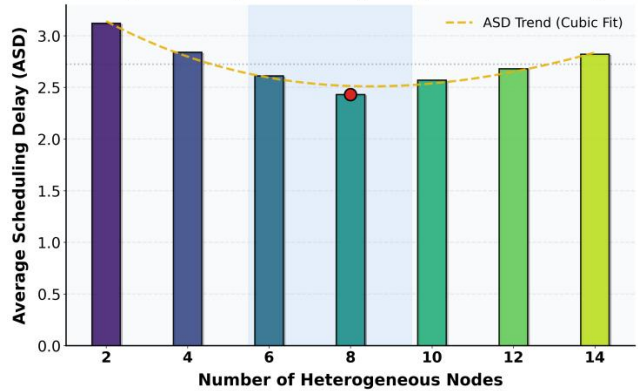

Figure 4. Sensitivity experiment on the average scheduling delay (ASD) of the heterogeneous node number change

As shown in Figure 4, the average scheduling delay (ASD) decreases first and then increases as the number of heterogeneous nodes grows, presenting a typical U-shaped trend. When the number of nodes is small (2-6), system resources are insufficient, and tasks are more likely to queue across nodes, resulting in higher overall scheduling delays. The performance bottleneck at this stage mainly comes from competition in computation and data transmission among nodes. This indicates that under low parallelism, heterogeneity does not effectively exploit the advantages of distributed processing. As the number of nodes increases to a moderate scale (around eight), ASD reaches its minimum, suggesting that resource parallelism and task granularity achieve an optimal balance, significantly improving task throughput and scheduling efficiency.

When the number of heterogeneous nodes increases further (beyond ten), ASD begins to rise again. This trend shows that excessive distribution introduces additional scheduling and communication overhead. The cost of state synchronization, load balancing, and data consistency maintenance among nodes gradually increases, leading to extra delay during task allocation and recovery. This indicates that in complex heterogeneous ETL environments, the relationship between node count and system performance is not linear. Too many nodes may even harm the stability of resource coordination. The experimental results verify that the proposed deep Q-learning-based scheduling framework exhibits strong adaptability under dynamic node changes. It maintains favorable scheduling delay performance as the scale of heterogeneous resources expands, demonstrating robustness and generalization advantages in multi-node task allocation and global scheduling optimization.

## IV. Conclusion

This study focuses on the intelligent ETL scheduling optimization problem in heterogeneous data environments and proposes a dynamic adaptive scheduling framework based on deep Q-learning. The framework models the complex ETL process as a Markov Decision Process in reinforcement learning, enabling the learning of optimal scheduling strategies under multi-source data, dynamic task dependencies, and heterogeneous resource constraints. By introducing a state representation network and a multi-objective reward function, the model balances key performance metrics such as delay, throughput, and resource utilization. It shows stable convergence and strong generalization in dynamic environments. Compared with traditional static or heuristic scheduling methods, this study provides new theoretical support and a practical pathway for building efficient and autonomous data pipelines, further promoting the deep integration of data engineering and reinforcement learning.

Looking ahead, the proposed framework can be extended and applied to a wider range of intelligent computing scenarios. For example, in high-concurrency environments such as cloud-native data warehouses, streaming data governance, and cross-region task orchestration, deep Q-learning can be combined with graph neural networks, meta-reinforcement learning, or self-supervised representation learning to enhance adaptability to task topology changes and environmental drift. Moreover, as computational scheduling and energy efficiency become key issues in data centers, future research can incorporate energy constraints, cost optimization, and multi-agent collaboration to achieve global scheduling decisions across clusters. Overall, this study not only provides a scalable intelligent scheduling paradigm for heterogeneous ETL systems but also lays a theoretical foundation for the algorithmic, autonomous, and interpretable development of intelligent data infrastructure.